\newif\ifarxiv

\arxivtrue

\ifarxiv
    \documentclass{article}
    \usepackage{arxiv}
    \usepackage[utf8]{inputenc} 
    \usepackage[T1]{fontenc}    
    \usepackage{hyperref}       
    \usepackage{url}            
    \usepackage{booktabs}       
    \usepackage{amsfonts}       
    \usepackage{nicefrac}       
    \usepackage{microtype}      
    \usepackage{amsmath}
    
    \newcommand\mdoubleplus{\mathbin{+\mkern-10mu+}}
    \usepackage{lipsum}
    \usepackage{graphicx}
    \usepackage[norule,symbol,perpage]{footmisc}

        \usepackage{caption}
    \usepackage{subcaption}
        \usepackage{bm} 
      \graphicspath{{imgs/}}
\else
     \documentclass[10pt,a4paper,conference]{IEEEtran}
    \usepackage{amsmath}
    \usepackage{booktabs}
    \usepackage{caption}
    \usepackage{subcaption}
    \usepackage{multirow}
    \usepackage{soul,color}
    \usepackage[utf8x]{inputenc}
    \usepackage{etoolbox}
    \usepackage{amssymb}
    \usepackage{breqn}
    
    \newcommand\mdoubleplus{\mathbin{+\mkern-10mu+}}
    \usepackage{bm}

\fi

\begin{document}
%

\title{Planar 3D Transfer Learning for End to End Unimodal MRI Unbalanced Data Segmentation}


\ifarxiv
\author{
    Martin Kolarik, Radim Burget\\
Dept. of Telecommunications\\
Brno University of Technology\\
Brno, Czech Republic\\
\texttt{martin.kolarik@vutbr.cz}
   \And
Carlos M. Travieso-Gonzalez \\
Dept. Signals and Communications. IDeTIC\\
University of Las Palmas de Gran Canaria\\
Las Palmas, Spain\\
\texttt{carlos.travieso@ulpgc.es}
  \And
Jan Kocica \\
Masaryk University, \\
and University Hospital Brno\\
University Hospital Brno\\
Brno, Czech Republic\\
}
\fi
\maketitle

\begin{abstract}
We present a novel approach of 2D to 3D transfer learning based on mapping pre-trained 2D convolutional neural network weights into planar 3D kernels. The method is validated by the proposed planar 3D res-u-net network with encoder transferred from the 2D VGG-16, which is applied for a single-stage unbalanced 3D image data segmentation. In particular, we evaluate the method on the MICCAI 2016 MS lesion segmentation challenge dataset utilizing solely fluid-attenuated inversion recovery (FLAIR) sequence without brain extraction for training and inference to simulate real medical praxis. The planar 3D res-u-net network performed the best both in sensitivity and Dice score amongst end to end methods processing raw MRI scans and achieved comparable Dice score to a state-of-the-art unimodal not end to end approach. Complete source code was released under the open-source license, and this paper complies with the Machine learning reproducibility checklist. By implementing practical transfer learning for 3D data representation, we could segment heavily unbalanced data without selective sampling and achieved more reliable results using less training data in a single modality. From a medical perspective, the unimodal approach gives an advantage in real praxis as it does not require co-registration nor additional scanning time during an examination. Although modern medical imaging methods capture high-resolution 3D anatomy scans suitable for computer-aided detection system processing, deployment of automatic systems for interpretation of radiology imaging is still rather theoretical in many medical areas. Our work aims to bridge the gap by offering a solution for partial research questions.
\end{abstract}


%

\section{Introduction} \label{introduction}
\footnotemark[0]Major magnetic resonance imaging (MRI) computer vision datasets, such as MICCAI 2016 MS lesion segmentation challenge dataset (MSSEG 2016) \cite{dataset}, consist of multiple perfectly co-registered and pre-processed modalities. However, the reality is not usually so generous in terms of data availability. Gathering medical data is expensive, and the needed pre-processing steps are still not automated with sufficient reliability \cite{zhang2019linear}. As Johnson states in the documentation to the ITK toolkit for processing biomedical imaging: "It is not uncommon for the registration process to run for several minutes and still produce a useless result." \cite{johnson2015itk}. These reasons support our arguments for developing algorithms processing raw medical data without pre-processing in the end-to-end paradigm and using only a single modality for training and prediction. \footnotetext[0]{This paper has been accepted for publishing at the ICPR 2020 Conference}
The aforementioned conditions, however, create additional challenges. Processing raw MRI data means that the problem of an unbalanced dataset cannot be solved simply by undersampling the training set and exploiting only those samples containing regions of interest. Without undersampling, the loss functions for unbalanced data such as Focal loss \cite{lin2017focal} do not always converge during training due to the absence of positive class on a large portion of samples. Additionally, the natural 3D data representation of MRI scans results in larger sample sizes, which lowers sample weighting effectiveness. As a solution to these problems in 3D data processing, we propose the planar 3D transfer learning, which helps with processing unbalanced data and offers improved generalization capabilities and shorter training time.

\begin{figure}[!t]
\centering
    \begin{subfigure}[b]{0.46\linewidth}
         \centering
         \includegraphics[width=0.8\linewidth]{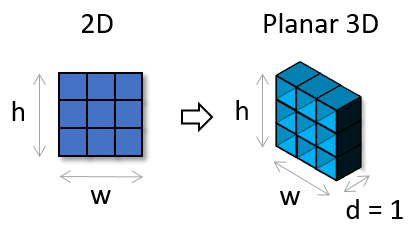}
         \label{fig:2dto3d}
         \caption{Convolutional kernel mapping from 2D to 3D \vspace{5pt}}
         \label{fig:planar3dkernel}
     \end{subfigure}
     \begin{subfigure}[b]{0.26\linewidth}
         \centering
         \includegraphics[width=0.86\linewidth]{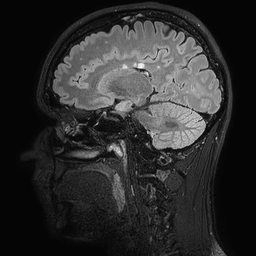}
         \caption{Raw MRI Sagittal scan}
         \label{fig:example_scan}
     \end{subfigure}
     \begin{subfigure}[b]{0.26\linewidth}
         \centering
         \includegraphics[width=0.86\linewidth]{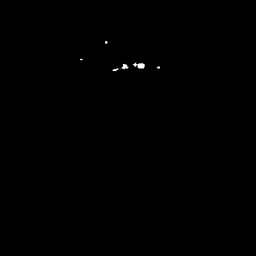}
         \caption{Segmentation mask}
         \label{fig:example_mask}
     \end{subfigure} 
\captionsetup{justification=centering}
        \caption{Visualisation of 2D $\mapsto$ Planar 3D convolutional kernel transformation and an example of MRI FLAIR sagittal scan slice with segmentation mask}
        \label{fig:datasetexample}
\end{figure}

The first problem we address with this paper is the need for transfer learning for 3D image analysis. While for 2D image processing, large labeled datasets exist (e.g., ImageNet \cite{imagenet}) and plenty of high quality pre-trained models are available, in the case of 3D imaging, the datasets are generally smaller, time for training grows rapidly, and 3D pre-trained high-quality models are missing. \\
Generally, the term "transfer learning" refers to a situation where knowledge learned from data distribution P$_1$ is exploited to improve generalization in the second data distribution P$_2$ \cite{goodfellow2016deep}. If there is significantly more data sampled from the first distribution P$_1$, then the learned data representation can be utilized to learn the distribution P$_2$ from very few samples. Transfer learning in deep convolutional neural networks (CNNs) is achieved by transferring the learned weights of an artificial neural network trained on P$_1$ and fine-tuning the model on P$_2$. Even if both data distributions are fairly different (general photo images in ImageNet \cite{imagenet} compared to MRI brain scan as in Fig. \ref{fig:datasetexample}), low-level features of the image such as edges, visual shapes, and geometric changes are similar in both distributions. These shared visual attributes are usually utilized in transfer learning by transferring weights from networks trained on a large labeled dataset for classification of 2D RGB images (e.g., Imagenet \cite{imagenet}) for processing of different datasets or tasks such as image segmentation \cite{albunet}. We propose a straightforward solution for weights transfer from 2D into 3D using planar 3D kernels and efficiently transfer existing 2D CNN architectures into the 3D image processing domain. Illustration of the planar transfer learning is in Fig. \ref{fig:planar3dkernel}. \\
The second issue addressed by this paper is the problem of segmenting heavily unbalanced data in 3D. Multiple sclerosis lesion segmentation is a typical example - segmented lesions make no more than 0.2 percent of voxels of the whole volume (see Fig. \ref{fig:datasetexample} (b) and (c) and Table \ref{tab:datasetscanner}), which makes the training extremely challenging. Not only do we lack training data in general due to the dataset small size, but also the regions of interest are critically underrepresented in the training data. In this work, we show how the problem of heavily unbalanced data segmentation can be solved by the implementation of planar 3D residual U-Net with encoder transferred from 2D VGG-16 \cite{vgg16} evaluating the performance on the MICCAI 2016 MS lesion segmentation challenge (MSSEG 2016) dataset \cite{dataset}. \\
The rest of the paper is structured as follows. In Section \ref{relatedwork}, we provide state-of-the-art in the area of transfer learning for segmentation and transfer learning for 2D to 3D. Section \ref{methodologydata} describes the experiment, the data preparation process, and all details needed to replicate the experiment, including a link to the complete source code. Section \ref{results} provides results of the experiments while the next Section \ref{discussion} discusses achieved results. The last section \ref{conclusion} sums up the main paper outcomes and states the vision of how we plan to continue with this research area.

\section{Related work} \label{relatedwork}

\begin{figure}[!t]
\centering
     \begin{subfigure}[b]{0.33\linewidth}
         \centering
         \includegraphics[width=1\linewidth]{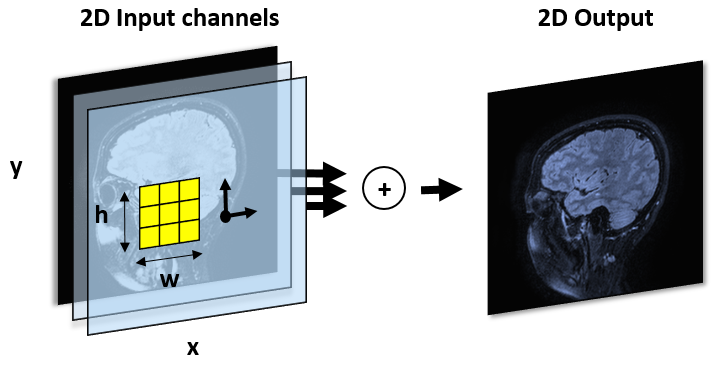}
         \caption{2D Convolution \vspace{10pt}}
         \label{fig:conv_2d}
     \end{subfigure}
     \begin{subfigure}[b]{0.33\linewidth}
         \centering
         \includegraphics[width=1\linewidth]{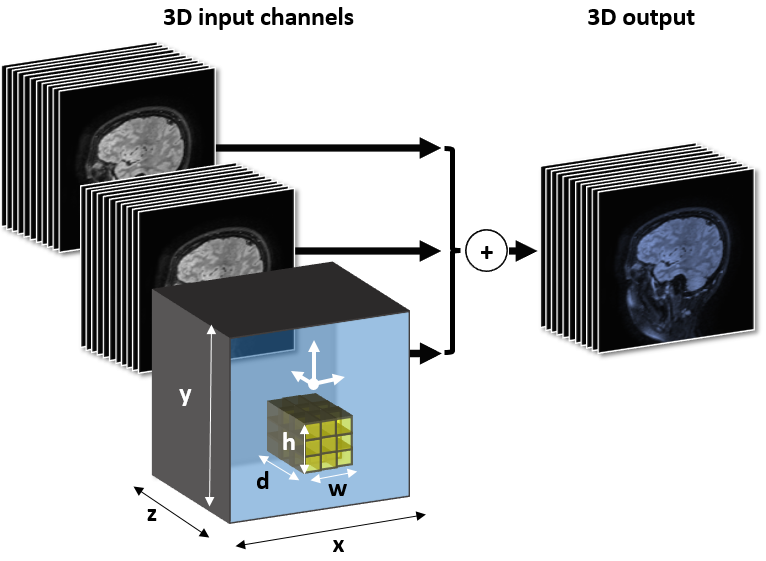}
         \caption{3D Convolution \vspace{10pt}}
         \label{fig:conv_3d}
     \end{subfigure}
     \begin{subfigure}[b]{0.33\linewidth}
         \centering
         \includegraphics[width=1\linewidth]{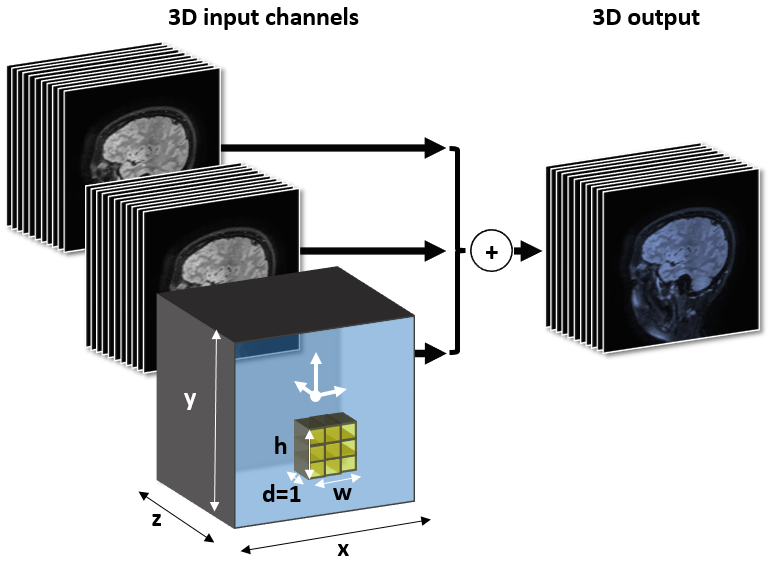}
         \caption{Planar 3D Convolution \vspace{10pt}}
         \label{fig:conv_pseudo_3d}%
     \end{subfigure}
     \captionsetup{justification=centering}
        \caption{Visualization of 2D, 3D, and planar 3D convolution operation. Notice that the difference between a 3D and a planar 3D convolution lies in the kernel depth.}
        \label{fig:convolution_types}
\end{figure}

Contemporary state-of-the-art techniques for semantic segmentation are almost exclusively based on deep CNNs \cite{taghanaki2019deep} and autoencoder network U-Net \cite{ Unet_original} inspires many of the successful architectures. Deep learning segmentation algorithms have proven its effectiveness in segmentation tasks of both outdoor or indoor general imaging \cite{yuan2019object}, in satellite image processing \cite{khryashchev2018comparison} and in medical imaging in two dimensional \cite{cai2017improving}, three dimensional \cite{milletari2016v} and hybrid 2D/3D \cite{hybriddenseunet} domain.

\subsection{Planar 3D networks}
A common misconception exists amongst deep learning practitioners that the important difference between 2D and 3D CNNs lies in the convolutional core's extra dimension. However, the main advantage of the 3D CNNs is that they represent the data as 3D volumes throughout the whole network, while 2D networks represent the data as 2D matrices only. This attribute makes 3D CNNs suitable for medical image processing. See Fig. \ref{fig:convolution_types} for detailed illustration of 2D, 3D and planar 3D types of convolutions. \\
Planar 3D networks were introduced as a solution to create more lightweight CNNs for 3D data processing. The planar 3D networks are a subset of 3D CNNs. General 3D CNN can use the convolutional core of size N × M × \textbf{L} (see Fig. \ref{fig:convolution_types} (b)) while the planar 3D networks consist of cores with size N × M × \textbf{1} (see Fig. \ref{fig:convolution_types} (c)). Planar 3D network convolutional core is of size 1 on at least one $x$, $y$, or $z$ axis. Although this property has been adopted as an overfitting prevention method, we show in this paper that the attribute can be effectively used to transfer weights from 2D to 3D CNNs as the 2D convolutional core (2D matrix) equals to a 3D convolutional core (3D tensor) with at least one dimension of size 1. Planar and multi-planar approaches have been adopted in 3D image processing. Examples include tri-planar classification and research of weight tying \cite{kim2020triplanar} or multi-planar 3D breast segmentation \cite{piantadosi2020multi}.

\subsection{Transfer learning from 2D to 3D networks} \label{2dto3d}
The two most common areas using 3D image representation are the medical 3D scans and video processing \cite{ji20123d}. When a person processes 3D data, we utilize the information in all three dimensions (spatial time axis for video). When neurologists label multiple sclerosis lesions in a human brain scan, they look in adjacent image slices to determine whether the voxel is or is not a part of the lesion. Nevertheless, in computer vision, mainly due to the practical unavailability of transfer learning in 3D neural networks, many approaches trade the more natural 3D representation for higher achievable accuracy on 2D representation exploiting the practical availability of transfer learning sources. \\
There are two possible approaches for 3D image transfer learning. The first option is training 3D networks designed for transfer learning on existing 3D datasets \cite{chen2019med3d}. Second is utilizing already existing 2D networks and transforming them for 3D processing. Experiments with inflated 2D to 3D kernels either by copying 2D kernels along the 3rd axis \cite{schmarje20192d} or by padding 2D convolutional kernel by zeros along 3rd axis \cite{zeropaddingtransfer} show promising results. However, 3D convolutional filters that consist of multiple stacked 2D filters do not contain 3D features of the input image, only adjacent 2D features. This method is logical for video processing (2D image + time as the third dimension) as the slices in the spatio-temporal axis are always one 2D image after another. The idea of kernel inflation where we take 2D convolutional kernels of size N × N  and stack them in the third dimension to size N × N × N was studied in the work of Carreira, et al. \cite{carreira2017quo}. Padding kernel with zeros adds computational complexity without adding gained value. Other notable approaches of transfer learning from 2D to 3D networks include works of Liu, et al. \cite{liu20183d} and research of Yu, et al. \cite{yu2019thickened} using thickened 2D networks. \\
Parallel to our experiments, there has been published work by Yang, et al. \cite{yang2019reinventing} who address the problem of 2D to 3D transfer learning by application of ACS convolutions, which are technically tri-planar convolutions with tied weights processing the image in an axial, sagittal and coronal axis. The method achieved good results, but it is also more computationally and memory demanding compared to our uni-axial method. While the computational requirements are not crucial because we only need to consider the inference phase for the transferred network part, the memory demands are the limiting factor in our experiment setting. The popular transfer learning CNN architectures such as VGG-16 are not primarily designed to be memory efficient, and the tri-planar approach effectively results in three running parallel instances of such network in memory. This restricts the method to use either smaller data representation or a more lightweight network as a transfer learning source. Both factors are limiting the tri-planar method's usage as a general 2D to 3D transfer learning solution.

\subsection{Transfer learning in semantic segmentation}
Transfer learning is a key part of the success of neural networks in recent years, and it is an effective method for training (fine-tuning) segmentation neural networks. The most common approach for autoencoder networks is based on the principle of transferring weights of the encoder part (sometimes called backbone) from a classification network trained on the Imagenet dataset \cite{imagenet}. Popular examples include the Ternausnet \cite{ternausnet}, which uses the VGG network \cite{vgg16} as encoder, or the Albunet \cite{albunet}, which uses the Resnet architecture \cite{resnet} as the encoder part. A popular segmentation model library \cite{segmentation_models} generalizes this principle of using a pre-trained encoder followed by a fine-tuned decoder. The reason why transfer learning from the general image domain performs well for medical data processing is an active research area. Experimental results show effectiveness in situations when the medical dataset is small in size \cite{raghu2019transfusion}, which is also the case of the MSSEG16 dataset.

\subsection{Image processing of unbalanced and limited data}
Some common approaches on data level or algorithm level are used to handle the problem of class imbalance. Data level approaches include undersampling or oversampling of the samples of interest, resulting in a more balanced dataset \cite{small2017handling}. However, the popular undersampling technique, which is based on training only on samples containing the infrequent class, is unusable when training on raw data. In our case, the CNN needs to see the whole image to not create false positives in non-brain areas. The algorithm level approaches include class weighting for 2D imaging \cite{mckinley2016nabla}, specialized loss function for unbalanced data such as focal loss \cite{lin2017focal} or its combination with Dice coefficient \cite{wang2018focal}. The problem becomes more severe with the growing size of the processed sample. We used a large 3D sample size in our experiment, and the network did not converge without implementing a planar 3D transfer learned encoder.

\section{Methodology} \label{methodologydata}
This paper aims to propose an efficient method of transferring existing pre-trained 2D neural network weights trained on large datasets (e.g., ImageNet) into 3D networks. Results were evaluated using a public dataset for multiple sclerosis lesion segmentation, and the proposed methodology is compared to other approaches. In comparison with other works, we focused on the problem of end to end learning and unbalanced data segmentation.\\
The complete source code used in this experiment is published on Github \cite{github}. As the dataset is also publicly available \cite{dataset}, the following section includes all important information to keep our research reproducible according to the Machine learning Reproducibility Checklist \cite{pineau2020improving}.

\subsection{Dataset MICCAI 2016 MS lesion segmentation challenge}
We evaluated our method over the MSSEG 2016 dataset \cite{dataset}, which consists of public training and the private testing part. Since only the training part is released and the testing is kept private, all information about the dataset in this paper will be related only to the public training part. An example of an input scan image and corresponding segmentation mask can be seen in Fig. \ref{fig:datasetexample}. \\
The dataset consists of 15 MRI head scans of 15 different patients who have multiple sclerosis scanned in 3 different medical centers. Information about the dataset is summarized in Table \ref{tab:datasetscanner}.

\begin{table}[!h]
\renewcommand{\arraystretch}{1.3}
\caption{Dataset details}
\label{tab:datasetscanner}
\centering
\begin{tabular}{lccc}
\toprule
        Center&  Lesion voxel ratio &        Scans &    Axial Resolution\\
\midrule
		01 &   0.311 \%    &    5 & 144 × 512 × 512\\
		07 &   0.141 \%   &    5 & 128 × 224 × 256\\
		08 &   0.144 \%   &    5 & 261 × 336 × 336\\
\midrule
Total & 0.199 \% & 15 & \\
\bottomrule
\end{tabular}
\end{table}

The size and quantity of multiple sclerosis lesions present in each patient scan differ largely as a result of the various progression of the disease. As shown in Table \ref{tab:datasetscanner}, a set of scans from each center consists of differently unbalanced scans. Center 01 includes patients with more severe disease (i.e., more lesions), while center 7 includes patients with less significant lesions. The average age of patients was 45.3 years (±10.3 years), with 40\% of males and 60\% females.

\subsection{Data preparation process}
\begin{figure}[!t]
\centering
     \begin{subfigure}[b]{0.49\linewidth}
         \centering
         \includegraphics[width=0.79\linewidth]{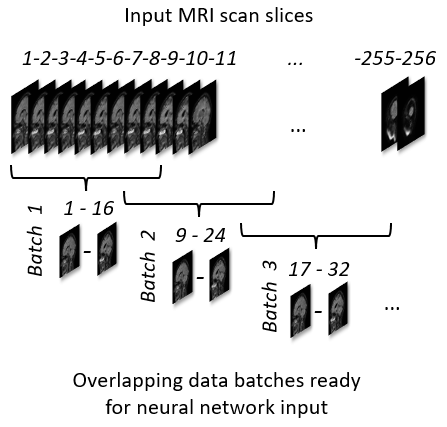}
         \caption{Preparation of overlapping data batches $B_{in}$ from the input MRI scan data set $S_{in}$}
         \label{fig:inputbatch}
         \vspace{10pt}
     \end{subfigure}
     \begin{subfigure}[b]{0.49\linewidth}
         \centering
         \includegraphics[width=0.79\linewidth]{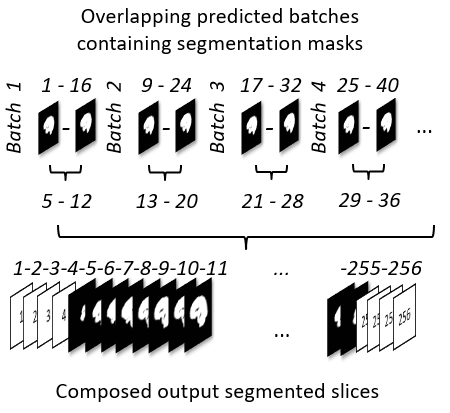}
         \caption{Composing the final output scan $S_{out}$ from inferred overlapping batches $B_{out}$}
         \label{fig:outputcompose}
         \vspace{10pt}
     \end{subfigure}
        \caption{Overlapping batches of training and testing data}
        \label{fig:datapreparation}
\end{figure}

To deploy our method into the medical praxis, we chose the unprocessed 3D FLAIR modality as a single data source input. We rotated the scans to sagittal view and rescaled them to the resolution of 256 × 256 × 256 using spline interpolation with spline order 2 (python package Scipy.ndimage.zoom). Before the training procedure, input scans have been normalized to values ranging [-1;1] and masks to values [0;1], because the output layer of the neural network uses sigmoid activation. Although we used VGG-16 pre-trained weights, which normally require input normalized between [-127, 127], we experimentally evaluated that experiments with an input data range of [-1;1] achieved better results. Because the VGG network is designed for RGB 3-channel input, we copied the single modality input into 3 channels.\\
After the pre-processing described in the previous paragraph, input scans $S_i$ with dimensions 256 × 256 × 256 were divided into overlapping batches $B_in$ of dimensionality 16 × 256 × 256 (see Fig. \ref{fig:datapreparation}). Therefore each $B_in$ consists of 16 sagittal slices with resolution 256 × 256 pixels. Batches $B_in$ were used as network input. Visualization of the batch preparation process can be seen in Fig. \ref{fig:datapreparation}. Corresponding batches of ground truth masks for training have been prepared using the same method. \\
We can describe the batch preparation process as dividing each input 3D scan $ S_in $, a set of 256 2D images, into subsets $B_in$ each containing 16 images using a sliding window with overlap 8. This process results in 31 batches of data $B_{in[1-31]}$ created from single scan $S_in$. The inferred output scan $S_out$ containing segmentation masks is then composed out of inferred output batches of data $B_{out[1-31]}$ as described in Equation \ref{eqDatacomposition} where symbol $\mdoubleplus$ stands for concatenation of two 3D matrices. The composition algorithm extracts the middle 8 slices from each overlapping batch (except the first $B_{out[1]}$ and last $B_{out[31]}$ from which we also extract the first 4 and last 4 slices respectively). By concatenating these extracted middle slices, we form the composed output scan $S_o$ containing 256 slices. Final results were obtained after normalizing the network predictions between [0;256] and thresholding them with the threshold value of 16 to obtain the final binary segmentation mask.

\begin{equation}
\label{eqDatacomposition}
\begin{aligned}
\bm{S}_{out} = {} & B_{out[1]}[1,12] \mdoubleplus B_{out[2]}[5,12] \mdoubleplus B_{out[3]}[5,12] \mdoubleplus \\
 & \mdoubleplus B_{out[4]}[5,12] \mdoubleplus \; ... \; \mdoubleplus B_{out[31]}[5,16]
\end{aligned}
\end{equation}

For evaluation, the 5-fold cross-validation was applied. During each round of the cross-validation, 1 scan from each center was selected for testing (in alphabetical order), which results in 5 rounds, as seen in Table \ref{tab:crossvalidation} after which we cross-validated the whole dataset. The rest of the 12 scans were used for training, from which 11 were used as a training set and the last 1 as a validation set. The training set was randomly shuffled before the training process.

\subsection{Planar 3D weights preparation}
The core idea of this paper lies in the mapping of 2D weights into 3D space $W_{2D}^{w × h} \mapsto W_{3D}^{w × h × 1}$. Weights of the convolutional layer consist of convolutional kernels and biases. Example 3D convolution operation of one convolutional layer $\beta$ is in Equation \ref{eqConvolution}. Here $\bm{Y}_\beta$ is the output of the convolutional layer $\beta$ which is summation of partial convolution operations (noted $*$) over the set of \textit{P} $\in \mathbb{N} \cap  [1, n]$ input 3D matrices (tensors) $\bm{X}_p$ and corresponding convolutional kernels $\bm{K}_\beta,_p$ with added scalar value bias $\mathit{B}_\beta$.

\begin{equation}
\label{eqConvolution}
\bm{Y}_\beta = \sum_{p=1}^{\mathit{n}} {(\bm{K}_\beta,_p * \bm{X}_p}) + \bm{B}_\beta
\end{equation}

Each convolutional layer weight has only single scalar value of bias B $\in \mathbb{R}$ (in case of tied biases). The mapping operation $W_{2D} \mapsto W_{3D}$ is then transcribed in Equation \ref{eqMap}:

\begin{equation}
\label{eqMap}
\bm{W}_{3D} : 
\begin{cases}
A_{i j \bm{1}} \in K_{3D} = A_{i j} \in K_{2D} \\
\mathit{B}_{3D} = \mathit{B}_{2D}
\end{cases}
\end{equation}

The operation depicted in Equation \ref{eqMap} transforms 2D convolutional kernels $K_{2D}$ of shape w × h to its planar 3D counterpart $K_{3D}$ of shape w × h × 1 by assigning the value of element $A_{i j} \in K_{2D}$ to the element $A_{i j 1} \in K_{3D}$ of planar 3D kernel.
Biases are kept as the same scalar value both in 2D and 3D representation. By reshaping the weights of the VGG-16 network according to Equation \ref{eqMap} we were able to effectively transfer the learned features into planar 3D processing as depicted in Fig. \ref{fig:convolution_types} and reuse them in the planar 3D VGG-16 backbone for 3D image processing.

\begin{figure*}[t]
\centering
\subfloat{\includegraphics[width=0.90\textwidth]{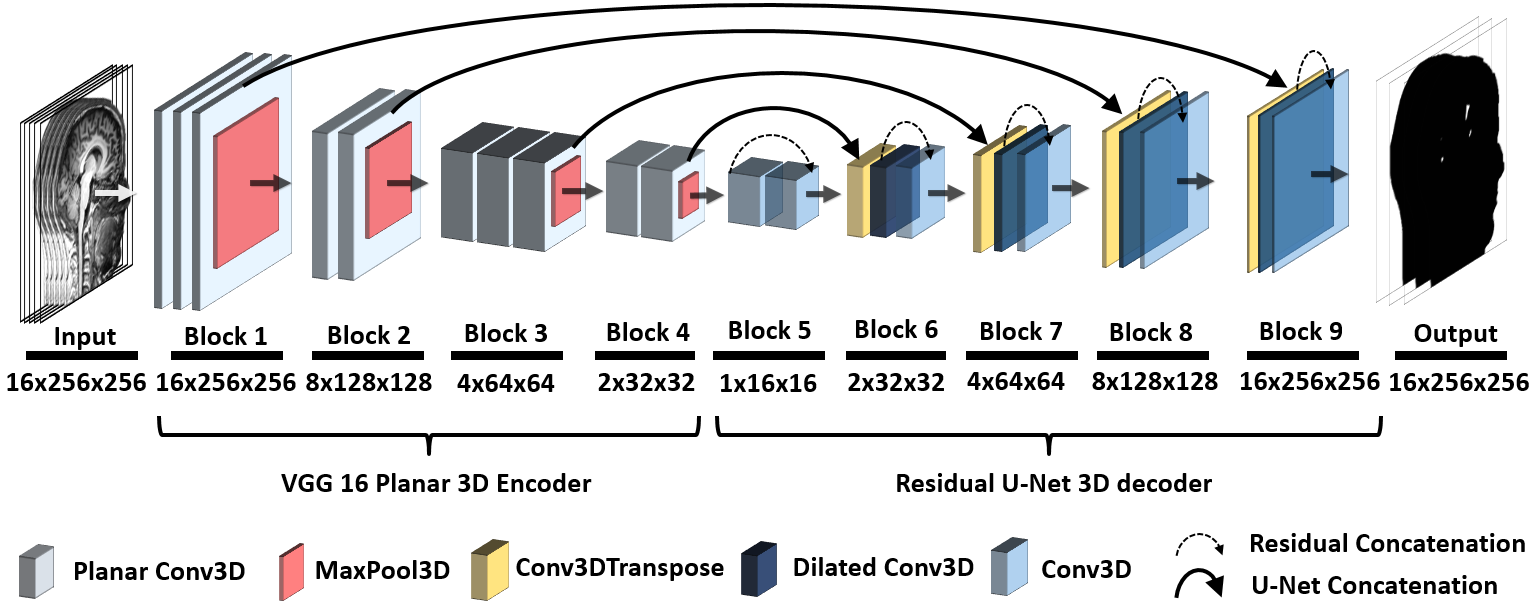}}
\captionsetup{justification=centering}
\caption{Architecture of the proposed planar 3D res-u-net used in this experiment. Encoder uses VGG-16 planar 3D transferred weights, while the decoder consists of standard 3D convolutional kernels and utilizes residual interconnections. \label{fig:architecture}}
\end{figure*}

\subsection{Neural network architecture}
The proposed architecture of planar 3D res-u-net is depicted in Fig. \ref{fig:architecture}. It consists of  planar 3D VGG-16 encoder \cite{vgg16} and decoder implementing residual U-Net from our previous research \cite{kolavrik2019optimized}. The 3D VGG-16 network was created from the original 2D VGG-16 by replacing the 2D convolutional and pooling layers with their 3D counterpart. \\
We have chosen the VGG-16 network as the backbone since it is easily transferable to planar 3D format and tested as a segmentation backbone for various medical image processing tasks. The decoder part is implemented as an u-net network \cite{Unet_original} with residual concatenation. Decoder block consists of the upsampling layer implemented by 3D transpose convolution followed by two 3D convolutional layers, of which the first uses dilated kernels \cite{yu2015multi} for a larger receptive field without additional parameter requirements. \\
The architecture does not contain batch normalization layers because 3D CNNs are highly memory demanding, leading to small batch size (in our case, the batch size was 2). Batch normalization can increase the network's error when the batch size is small due to the inaccurate batch statistics estimation \cite{wu2018group}. 

\subsection{Neural network training procedure} \label{nntrain}
We evaluated the performance of the proposed CNN trained under the different settings of hyperparameters during our experiments. During training, one of the critical problems was the neural network's ability to learn to segment such a heavily unbalanced dataset. We evaluated various loss functions starting with standard binary cross-entropy, dice coefficient, Tversky loss, and focal loss for unbalanced data \cite{abraham2019novel}. None of the tested loss functions was able to segment the data without transfer learning. The wrongly converged network predicted all voxels to be negative, achieved relatively high accuracy, and ended stuck in a local minimum of the loss function. We believe the transfer learning helped solve this issue because the transferred encoder posed as a static feature extractor for the decoder part during the training process, helping the network easier find the global minimum of the loss function.\\
Even with transfer learning, some settings of hyperparameters did not lead to convergence. To prevent this behavior, a small learning rate Lr $\leq$ $1e-4$ was adopted. We achieved the best results using Lr = $5e-5$. Optimizer Adam \cite{kingma2014adam} was used with default settings from library Keras and with value decay = $1.99e-7$. \\
We set the maximum training epochs to 100 with early stopping if the validation loss did not improve in the subsequent 40 epochs. Because of the low learning rate, the training process usually took all 100 epochs to improve validation loss gradually. \\

\begin{equation}
\label{eqDice}
D(X,Y) = \frac{2*\left|X \cap Y\right|}{\left|X\right|+\left|Y\right|} = \frac{2 * TP}{\left|X\right|+\left|Y\right|}
\end{equation}

The best results were achieved by a combination of binary cross-entropy and dice coefficient. See Equations \ref{eqDice} and \ref{eqLoss} for more details of the used loss function.

\begin{equation}
\label{eqLoss}
Loss(X,Y) = BC(X,Y) - D(X,Y) + 1
\end{equation}

\subsection{Implementation details} \label{implementation}
Implementation was done in Tensorflow \cite{tensorflow2015-whitepaper} using Keras library \cite{chollet2015}. Computation experiments were performed on GTX 1080ti and RTX 2080ti GPU cards with 11 Gb of memory. Depending on the chosen loss function and validation metrics, one epoch lasted between 100-120s, which resulted in 2.5-3.5 hours of training. The inference of a single 3D brain scan segmentation mask took approximately 15s.

\section{Results} \label{results}
During the evaluation of the results, we aimed to simulate the MSSEG 2016 challenge conditions as closely as possible without access to the test set. We used the 5-fold cross-validation scheme and calculated the chosen metrics (Dice coef. - Equation \ref{eqDice} and sensitivity - Equation \ref{eqSensitivity}) according to the methodology used in the challenge.

\begin{equation}
\label{eqSensitivity}
Sensitivity = Recall = \frac{TP}{TP + FN}
\end{equation}

Each metric was calculated separately for every testing scan. One testing scan per center was used every round of cross-validation selected by alphabetical order. We then calculated the mean and standard deviation of the Dice coef. and sensitivity across all scans. This evaluation scheme is important to ensure compliance with MSSEG 2016 challenge evaluation process. \\
The results of cross-validation steps can be seen in Table \ref{tab:crossvalidation}. It is important to notice how the algorithm performed on different scanning centers. The worst results in both metrics were obtained on the center containing the lowest number of positive voxels. Our final results compared to other works can be seen in Table \ref{tab:final_results}.

\begin{figure*}[ht]
\captionsetup{justification=centering}
\centering
     \begin{subfigure}[b]{0.195\textwidth}
         \centering
         \includegraphics[width=1\textwidth]{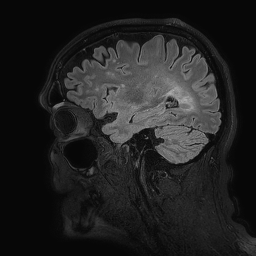}
         \caption{Raw MRI FLAIR scan slice (our input)}
         \label{fig:comparison_scan_raw}
     \end{subfigure}
    \begin{subfigure}[b]{0.195\textwidth}
         \centering
         \includegraphics[width=1\textwidth]{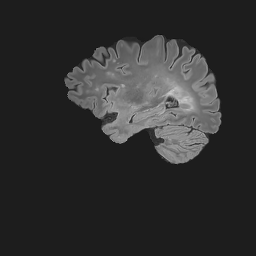}
         \caption{Pre-processed, brain extracted MRI}
         \label{fig:comparison_scan_preprocessed}
     \end{subfigure}
     \begin{subfigure}[b]{0.195\textwidth}
         \centering
         \includegraphics[width=1\textwidth]{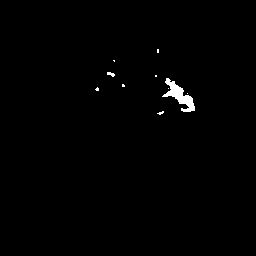}
         \caption{Mask predicted by our system}
         \label{fig:comparison_mask}
     \end{subfigure}
          \begin{subfigure}[b]{0.195\textwidth}
         \centering
         \includegraphics[width=1\textwidth]{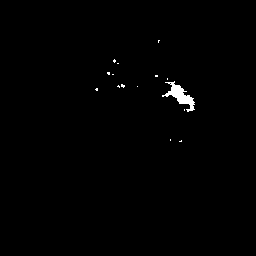}
         \caption{Consensus segmentation mask}
         \label{fig:comparison_consensus}
     \end{subfigure}
          \begin{subfigure}[b]{0.195\textwidth}
         \centering
         \includegraphics[width=1\textwidth]{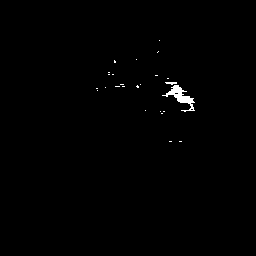}
         \caption{Mask labeled by one human expert}
         \label{fig:3d}
     \end{subfigure}
     \captionsetup{justification=centering}
        \caption{Examples of raw and pre-processed MRI FLAIR scans with corresponding mask predicted by our system compared to consensus mask and mask labeled by one of the experts. Consensus labels were merged from 7 different expert annotations.}
        \label{fig:comparison}
\end{figure*}

\section{Discussion} \label{discussion}
Our planar 3D res-u-net achieved a score of 0.611 for Dice coefficient outperforming the best published end to end method with a score of 0.512 and scoring within the confidence interval of the best unimodal not end to end method by McKinley, et al. \cite{mckinley2016nabla} with a score of 0.598 (winning method of the MSSEG challenge). Our method also gave the best result in sensitivity amongst end-to-end solutions scoring 0.60, exceeding the second approach processing raw MRI with a score of 0.45.

\begin{table}[!t]
\renewcommand{\arraystretch}{1.3}
\caption{Dice coefficient \& Sensitivity - 5-fold cross-validation results over each imaging center}
\label{tab:crossvalidation}
\centering
\begin{tabular}{lcccc}
\toprule
        \textbf{Metric}&          Center 1&              Center 7&               Center 8&  	 \textbf{Average}\\
\midrule
		\textbf{Dice} &      ${0.69 \pm 0.11}$&    ${0.50 \pm 0.11}$&     ${0.64 \pm 0.09}$&	${\mathbf{0.61 \pm 0.05}}$\\
        \textbf{Sens.} &  ${0.61 \pm 0.14}$&    ${0.45 \pm 0.13}$&     ${0.72 \pm 0.16}$&		${\mathbf{0.60 \pm 0.10}}$\\
\midrule
\end{tabular}
\end{table}

\begin{table}[!t]
\renewcommand{\arraystretch}{1.3}
\caption{Our final results compared with other unimodal methods and human expert results}
\label{tab:final_results}
\centering
\begin{tabular}{lccc}
\toprule
\multicolumn{4}{c}{Unimodal methods processing raw input data end-to-end}\\
        Authors  & End-to-end&          Dice coef. &      Sensitivity\\
\midrule
         Knight, et al. \cite{knight2016ms} &    Yes &   ${0.512 \pm 0.014}$    &  ${0.45 \pm 0.05}$\\
    \textbf{Our Result} &\textbf{Yes}&          		${\mathbf{0.611 \pm 0.052}}$   &  ${\mathbf{0.60 \pm 0.10}}$\\
\midrule
\multicolumn{4}{c}{Unimodal methods processing input data after brain extraction}\\
        Authors  & End-to-end&          Dice coef. &      Sensitivity\\
\midrule
         Mahbod, et al. \cite{mahbod2016automatic} &   No&   ${0.448 \pm 0.064}$   &  ${0.53 \pm 0.03}$\\
         McKinley, et al. \cite{mckinley2016nabla}&    No& 	${0.598 \pm 0.059}$   &  ${0.65 \pm 0.03}$\\
\midrule
\multicolumn{4}{c}{Human expert results}\\
        \multicolumn{2}{l}{Expert annotation}&          Dice coef. &      Sensitivity\\
\midrule
         \multicolumn{2}{l}{Least precise human expert}&            		${0.670 \pm 0.008}$   &  ${0.72 \pm 0.06}$\\
        \multicolumn{2}{l}{Average human expert} &          ${0.705 \pm 0.027}$  &  ${0.77 \pm 0.02}$\\
\bottomrule
\end{tabular}
\end{table}

Compared methods by Mahbod, et al. \cite{mahbod2016automatic} and McKinley, et al. \cite{mckinley2016nabla} use single modality input, but unlike our method, they use the pre-processed version of the input data (skull stripped, brain extracted, normalized). Even though these approaches do not process data in an end to end fashion and therefore are not directly comparable to our results, we decided to include them for comparison as we did not find other published unimodal methods evaluated over the MSSEG 2016 dataset (neither more recent nor from the original challenge). Although the unimodal approach is also promoted by the organizers of the MSSEG 2016 challenge, it has not been adopted by the majority of other published methods \cite{hashemi2018asymmetric}.\\
Additionally, our results show the effectiveness of the approach both in transfer learning and unbalanced data segmentation. Because the explainability and transfer learning from general to medical image domains is currently a subject of research, the reasoning why the method achieved the published results is limited. As the results in \cite{raghu2019transfusion} show, there is a measured improvement when knowledge is transferred from a large to a small image dataset. This is the case of our study due to the small size of the MSSEG16 dataset (15 MRI scans) and the fact that the area of interest (lesions) is critically underrepresented in the segmented data. The effect of transfer learning on unbalanced data segmentation and the network's ability to converge to a global minimum of the loss function is the subsequent area of our research. We think this behavior happened due to the fixed encoder posing as a feature extractor with weights trained on a larger dataset. \\
In contrast to other methods implementing deep CNNs \cite{mahbod2016automatic}  \cite{mckinley2016nabla}, which use multi-stage detectors and multiple networks for prediction, our method uses a single-stage approach utilizing a single neural network. This helps to retain relatively low both training and inference time, as mentioned in Section \ref{implementation}. The inference time of 15s per one MRI scan shows the practical usability of the method in daily medical praxis. The memory requirements of our method are also lower when compared to the tri-planar method of 2D to 3D transfer learning \cite{yang2019reinventing}, which enables us to transfer memory inefficient architectures such as VGG-16.

\section{Conclusion and future work} \label{conclusion}
We presented a novel end-to-end learning approach to transfer learning utilizing mapping existing CNNs weights from 2D to planar 3D representation. This enables the 3D CNN image processing methods to keep pace with the more rapid development of 2D methods. We have shown the effectiveness of our approach on the problem of heavily unbalanced data semantic segmentation, where we were able to successfully segment multiple sclerosis lesions from raw MRI unimodal scans without selective sampling, class, or sample weighting. All our experiments that did not use the planar 3D transfer learning failed in segmenting the multiple sclerosis lesions converging to an incorrect local minimum of loss function predicting all voxels negative. Additionally, our approach operates with single modality input, which is an advantage in real praxis as it does not require co-registration nor additional scanning time during an examination. \\
Our method performed the best both in sensitivity and Dice coefficient amongst end to end methods processing raw MRI data without brain extraction and achieved comparable Dice coef. to a state-of-the-art unimodal not end to end approach. Thanks to the transfer learning, we significantly reduced training time to an average of 3 hours. Since the proposed method uses an end to end learning approach, it can also be directly applied to many other segmentation tasks without the need for manual pre-processing, co-registration, or additional modalities. Our planar 3D transfer learning method can be utilized for general 2D to 3D transfer learning between existing architectures. We plan to test it thoroughly on other 3D biomedical image domain processing tasks and other CNN architectures. \\
We plan to focus on further improvement of the unimodal raw MRI image processing method for multiple sclerosis lesion segmentation in the future. The next steps will include merging predictions from overlapping output batches, research of explainability of convergence in an unbalanced data environment, and testing equivariant roto-translation networks with planar 3D transferred weights.

\section*{Acknowledgment}
Research described in this paper was supported by the MPO FV20044, National Sustainability Program under grant LO1401 and by European Regional Development Fund, project Interreg, niCE-life, CE1581. Supported by Ministry of Health of the Czech Republic, grant nr. NV18-08-00459. All rights reserved. Specific Research project ref. MUNI/A/1325/2019 provided by Masaryk University Brno, and by the Ministry of Health of the Czech Republic project for conceptual development in research organizations, ref. no. 65269705 (University Hospital Brno, Brno, Czech Republic).

\newpage
\bibliographystyle{IEEEtran}
\bibliography{root.bib}

\end{document}